\documentclass{article}

\PassOptionsToPackage{round, comma, sort&compress, authoryear}{natbib}
\usepackage[preprint]{neurips_2025}
\usepackage[T1]{fontenc}
\usepackage[utf8]{inputenc}
\usepackage{microtype}
\usepackage{textcomp}

\PassOptionsToPackage{hyphens}{url}
\usepackage{url}
\usepackage{amsmath}
\usepackage{amssymb}

\usepackage{booktabs}
\usepackage{array}
\usepackage{longtable}
\usepackage{calc}              
\providecommand{\real}[1]{#1}
\usepackage{enumitem}

\usepackage[font=small,labelfont=bf]{caption}

\usepackage{graphicx}
\graphicspath{{figures/}}

\usepackage{xcolor}

\usepackage[colorlinks=true,
            linkcolor=blue!50!black,
            citecolor=blue!50!black,
            urlcolor=blue!50!black]{hyperref}
\usepackage[capitalize, noabbrev]{cleveref}

\title{Calibrating Interpretability Instruments Before Trusting Their Verdicts}

\author{%
  Orion Reblitz-Richardson\thanks{%
    Distiller Labs. Correspondence to Distiller Labs
    \textless\texttt{orion@orionr.com}\textgreater.}
}

\date{August 2026}

\providecommand{\tightlist}{%
  \setlength{\itemsep}{0pt}\setlength{\parskip}{0pt}}

\usepackage{fancyvrb}
\DefineVerbatimEnvironment{Highlighting}{Verbatim}{commandchars=\\\{\}}

\begin{document}

\maketitle

\begin{abstract}
Causal claims about large language model (LLM) internals rest on measurements. Those might
include a projection, a cosine, an ablation delta, or an interchange patch among others. These
measurements fail in specific, diagnosable ways that return a plausible number instead of an
error, so a broken instrument can easily read as a finding. A covariance-matched null can
saturate until every direction looks typical, a per-head attribution can overshoot the true
residual write threefold on reordered-normalization architectures, an interchange patch can go
sign-chaotic because its outcome is pinned at a ceiling, or a read-from verdict can be an
artifact of measuring past the layer where the model already decided.

This note documents six such failures from a causal interpretability program on refusal and
moral representation, spanning several papers and a four-model open-weight panel; each mode is
established on one or two of the four. For each we give the tell that catches it and a protocol
keyed to a detectable trigger (reordered normalization, massive activations, a low-dimensional
decision channel), so we and readers can check whether a given setup is exposed. The discipline
reduces to four moves: calibrate against a positive-control ladder, certify with an orthogonal
cell, compute power before spending compute, and state every read-from verdict at a depth
referenced to the model's commitment. The evidence is four architectures across three families
within a single program; external replication across programs is future work.
\end{abstract}

\section{Introduction}\label{introduction}

A causal interpretability program kept
producing results that dissolved under an instrument check. Each of the failures below
looked like a finding first. This note collects the six that turned into portable methods
findings and the estimator and intervention patterns the program
re-derived, and states each as a protocol we found portable within this program and offer
for others to test. The scientific results
(what refusal reads, how it commits) live in the companion flagship paper
\citep{reblitzrichardson2026slice};
this note is the portable methodology. Numbers here trace to the program's claim ledger;
every scalar carries its detection bar or its control.

The discipline is four moves: \textbf{calibrate the instrument against a positive-control ladder,
certify it with an orthogonal cell, compute power before spending compute, and state every
read-from verdict at a stated depth relative to the model's commitment.} Each section
below takes one instrument, shows the failure as it first appeared, names the tell that
caught it, gives the protocol, and states the check that certifies the fix.

\textbf{Notation and definitions.} The \emph{participation ratio} of a position is
\(\mathrm{PR} = (\sum_i \lambda_i)^2 / \sum_i \lambda_i^2\), where the \(\lambda_i\) are the
eigenvalues of the residual-stream covariance at that position; it is an effective-dimension
estimate of that covariance, equal to the true dimension for an isotropic covariance and
dropping toward 1 as variance concentrates in a few directions. The \emph{decision site} (or
control-token position) is the assistant-header or end-of-prompt control token where the
refusal gate and the judgment direction are defined; \emph{content positions} are the token
positions carrying the request text. The \emph{moral subspace} is the rank-3 span of the moral
mean-difference directions extracted from the moral-content datasets. The \emph{refusal-decision}
and \emph{judgment-decision} directions are the mean-difference directions for refuse-versus-comply
and for the moral judgment, read at the decision site. The \emph{reconstruction fractions}
\(R_{\mathrm{refusal}}\) and \(R_{\mathrm{judgment}}\) (written \path|R_refusal| and \path|R_judgment| below)
give the share of the full interchange-patch effect on the refusal (respectively judgment)
outcome that a restricted rank-\(k\) subspace patch transfers, normalized to \([0,1]\), so \(0\) is
no transfer and \(1\) is full reconstruction. \emph{Engage} and \emph{disengage} name the two
directions of a content intervention: engage adds harmful content, disengage removes it.
Terms are defined again at first use below.

The six failure modes, with the section that treats each:

\begin{enumerate}
\def\labelenumi{\arabic{enumi}.}
\tightlist
\item
  A projection-fraction instrument reads absence at a position where its own positive control
  has no discriminating power (the band-below-null decision site; \Cref{a2-band-below-null}).
\item
  A massive-activation outlier is a content-position statistic, so the decision-token
  bottleneck it seemed to contaminate is in fact clean (\Cref{a5-outlier}).
\item
  A covariance-matched null saturates in massive-activation families until every direction
  projects like a typical one (\Cref{a1-covariance-null}).
\item
  A per-head OV attribution overshoots the true residual write about threefold on
  reordered-norm architectures (\Cref{a3-ov-attribution}).
\item
  A deliberation/prefill asymmetry statistic is operating-point-confounded when one arm sits
  at the ceiling (\Cref{a6-deliberation}).
\item
  A cross-model asymmetry measured at the read layer is an artifact of measuring past the
  layer where one model already committed (\Cref{depth-discipline}).
\end{enumerate}

Coverage is uneven across the panel: each mode is established on one or two models, not on all
four, and the per-mode breakdown is in the Limitations section. Table 1 collects the
participation ratios by model, position, and normalization so that a reader can tell which PR
belongs where.

Mechanistic claims about model internals are claims about measurements. ``Refusal reads the
harm percept,'' ``this head writes the refusal direction,'' ``moral judgment is orthogonal to
the refusal decision'': each is a statement about a projection, a cosine, an ablation delta,
or an interchange patch. The measurement can fail in ways that produce a clean-looking
number. A covariance-matched null can saturate so that every direction projects like a
typical one. A per-head attribution can overshoot the true residual write threefold. An
interchange patch can go sign-chaotic because the outcome it reads is pinned at its ceiling.
A ``reads-X'' verdict can be an artifact of measuring past the layer where the model already
decided. None of these announce themselves; each returns a plausible scalar.

The discipline in this note was forced by one discovery. Across four architectures, the
decision site (the assistant-header or end-of-prompt control token where the refusal gate
and the judgment direction are defined) is a low-dimensional bottleneck. The participation
ratio there is 14.7 on OLMo-3-7B-Instruct, 8.6 on Qwen2.5-7B, 10.2 on Llama-3.1-8B, and
12.8 on GPT-OSS-20B (a 20B reasoning MoE at its harmony decision token). A 9-to-15
effective-dimensional channel, on every model tested, while content positions at the same
layers are full-rank-healthy (PR 40+/33+/35+). This is a substantive finding about where
the refusal decision lives, and it belongs in the flagship \citep{reblitzrichardson2026slice}. But it is also the reason the
program's projection-fraction instruments failed: a positive control measured in a 15-slot
channel projects onto its own span \emph{less} than a random direction does, so the instrument
had no discriminating power exactly where the interesting directions live. The finding and
the failure are the same fact seen twice. This note carries the validity protocol the
finding motivated; the flagship carries the finding.

\begin{figure}[t]
\centering
\includegraphics[width=\linewidth]{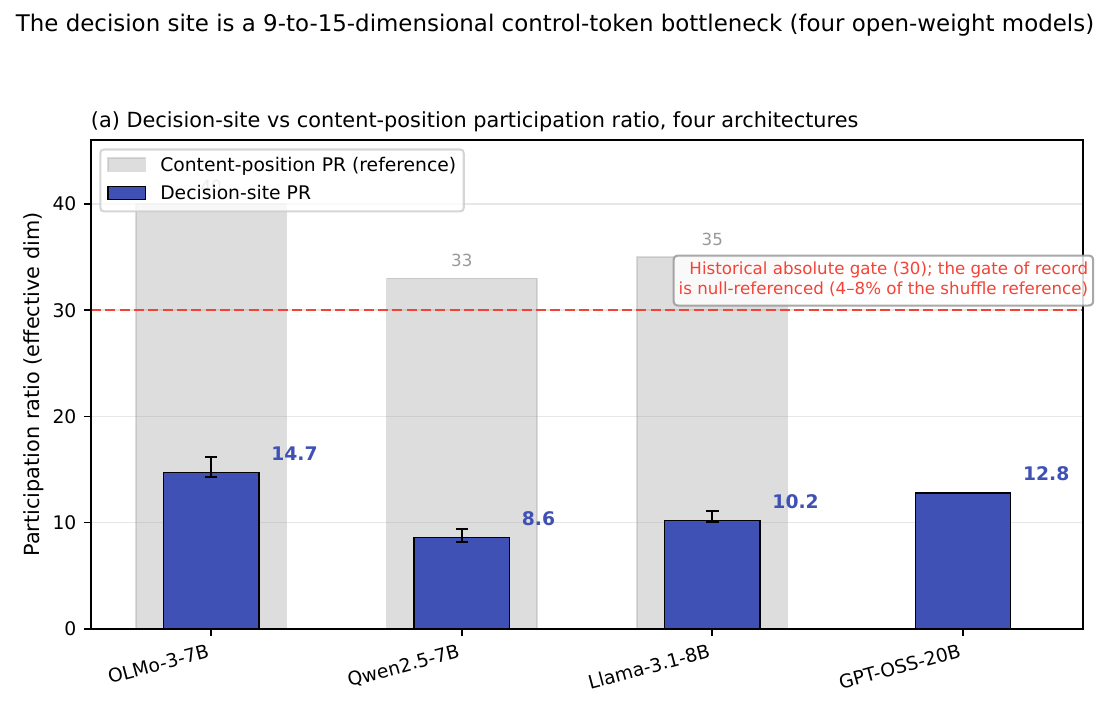}
\caption{The decision-site participation ratio across four architectures: OLMo-3-7B-Instruct 14.7, Qwen2.5-7B 8.6, Llama-3.1-8B 10.2, and GPT-OSS-20B 12.8 (a 20B reasoning MoE at its harmony decision token). Whiskers are 95\% subsampling intervals on one 240-text sample per model (128 texts for GPT-OSS): OLMo [14.3, 16.2], Qwen [8.2, 9.4], Llama [10.1, 11.1] around its value of record 10.2 (10.3 on that sample); GPT-OSS is plotted at its standardized 12.8 from the Tier-1 session (raw 9.4 [9.1, 10.7] on the W4 sample). The faint reference bars are the in-format-ladder content-position values of record (40 / 33 / 35); Table 1 lists the W4 sample's content positions. All four sit at 4 to 8 percent of the column-shuffle reference (the historical absolute gate of 30 is drawn for orientation), while content positions at the same layers stay full-rank-healthy (PR 40+/33+/35+). The refusal decision lives in a 9-to-15 effective-dimensional control-token channel, on every model tested.}
\label{fig:bottleneck-pr}
\end{figure}

(The bottleneck PR bar across the four architectures is \textbf{Figure~\ref{fig:bottleneck-pr}}, which uses the raw
in-format-ladder values with subsampling intervals: OLMo 14.7 {[}14.3, 16.2{]}, Qwen 8.6 {[}8.2, 9.4{]},
Llama 10.2 of record (10.3 {[}10.1, 11.1{]} on the later 240-text sample), GPT-OSS 9.4 {[}9.1, 10.7{]}
raw and 12.8 standardized from the Tier-1 session. The Llama 13.5 quoted in \Cref{power-tables} is the
decision-anatomy harness's standardized read of the same position on a different sample. Every
value is 4 to 8 percent of its column-shuffle reference.)

\section{Related work}\label{related-work}

Each failure mode in this note has roots in an established line of work. We state the connection
and what this note adds, so the reader can judge which cautions are new instruments and which are
new framings of known ones.

\textbf{Outlier dimensions and massive activations.} That a few residual-stream dimensions carry a
disproportionate share of variance, and that they distort geometric measurements, is well
documented: rogue dimensions obscure representational quality under cosine similarity
\citep{timkey2021rogue}, a small set of outlier dimensions disrupts transformers when removed
\citep{kovaleva2021bert}, and emergent outlier features appear at scale
\citep{dettmers2022int8, sun2024massive}, related to the attention-sink phenomenon
\citep{xiao2023efficient}. The standard response, per-dimension standardization, is not new
here. What we add is that the \emph{covariance-matched null} built for projection-fraction tests
silently degenerates in these families: because the null draws random directions from the
outlier-dominated covariance, every direction projects like a typical one, so the test loses
discriminating power rather than returning an obvious artifact. Naming this instrument-level
failure, and the band-below-null tell that catches it, is the contribution.

\textbf{Norm handling in attribution.} Reading a direction's per-head or per-layer contribution off
the residual stream requires accounting for the block's normalization; the logit-lens and
tuned-lens line makes the sensitivity to that choice explicit \citep{belrose2023tunedlens}.
Folding the block RMSNorm gain into per-head attribution is itself standard interpretability
tooling \citep{elhage2021framework, nanda2022transformerlens}, so the fold is not the contribution.
What we add is the specific, quantified failure for reordered (post-block) normalization as used
by the OLMo-2/3 family: a naive per-head decomposition that skips the block norm overshoots the
true residual write about threefold, which we quantify and catch with a two-sided reconstruction
gate (a one-sided floor misses overshoot).

\textbf{Activation-patching methodology.} That patching verdicts depend on metric, corruption, and
layer choice is the subject of best-practice work \citep{zhang2024patching}. Our stimulus and
depth sections are instances. An interchange readout run at a saturated outcome yields
sign-chaotic deltas that mimic instrument failure, and a read-from verdict measured past the
layer where the model has already committed can be a read-layer artifact. We frame both for
reasoning models and supply the orthogonal-cell certificate and the commitment-relative depth as
the checks.

\textbf{Interpretability illusions.} The general hazard, that a measurement can behave as if it found
a mechanism when it has not, is the illusion literature: subspace activation patching can route
through unintended pathways \citep{makelov2023subspace}, and individual-unit interpretations can
be spurious \citep{bolukbasi2021illusion}. This note's thesis, that a broken instrument reads as
a finding, is in that spirit. Our addition is the position-validity check, where a
positive-control band that falls below the covariance null marks the measurement position itself
as uninformative, and the integration of these checks into a pre-registration and verification
protocol.

The scientific results that exercise these instruments are reported in the companion flagship paper
\citep{reblitzrichardson2026slice}; this note is the portable methodology, and its evidence is the model panel and
single program it was derived on (see the limitations).

\section{The decision-site instrument and its calibration ladder}\label{decision-site}

Four failures converge on one object: the projection-fraction / cosine instrument used to
ask whether a direction of interest lives inside a subspace. One is the position where it
fails, another shows that position is architecture-general, a third is the null that degenerates
underneath it, and a fourth is the attribution decomposition that overshoots when the same
channel is read per-head. Each is stated as: failure $\to$ tell $\to$ protocol $\to$ certifying check.

Several participation ratios recur below, at different positions and under different
normalizations. Table 1 lists them together so that each PR can be traced to its model,
position, and normalization.

\begin{longtable}[]{@{}
  >{\raggedright\arraybackslash}p{(\linewidth - 10\tabcolsep) * \real{0.1667}}
  >{\raggedright\arraybackslash}p{(\linewidth - 10\tabcolsep) * \real{0.1667}}
  >{\raggedright\arraybackslash}p{(\linewidth - 10\tabcolsep) * \real{0.1667}}
  >{\raggedright\arraybackslash}p{(\linewidth - 10\tabcolsep) * \real{0.1667}}
  >{\raggedright\arraybackslash}p{(\linewidth - 10\tabcolsep) * \real{0.1667}}
  >{\raggedright\arraybackslash}p{(\linewidth - 10\tabcolsep) * \real{0.1667}}@{}}
\caption{Participation ratio (PR = ($\Sigma$$\lambda$)$^2$/$\Sigma$$\lambda$$^2$) by model, position, and normalization, measured on one
240-text sample per model (128 for GPT-OSS) at the primary layer. The raw decision-site column is
the value plotted in Figure 1; intervals are subsampling intervals over texts (the row-resampling
bootstrap is biased low because duplicated rows lower the sample rank, and is not used). The
standardized column is the same position after per-dimension z-scoring. The Llama 13.5 quoted in
\Cref{power-tables} comes from the decision-anatomy harness (standardized, request-twin stimuli, a different
sample): one position under a second harness and normalization, not a second token; the two
standardized reads (13.5 there, 14.2 here) agree to within 0.7, and the in-format value of record
stays 10.2 (10.3 on this sample). Content-position PRs are full-rank-healthy. The shuffle reference is the PR
after independent column permutations, which keeps every marginal variance and destroys the
correlations: every decision site sits at 4 to 8 percent of it, and at 3.6 to 7.4 percent of its
sample-rank ceiling, while the content positions of the same texts sit 2.6 to 9.4 times higher.
The geometric-cell column is the raw $\to$ standardized pair of \Cref{a1-covariance-null}, where per-dimension
standardization lifts Qwen and Llama out of near-rank-1 collapse. GPT-OSS 12.8 is its harmony
decision-token PR, treated as position-valid for the refusal decision-direction read against a
separate MoE PR ceiling of 25 (\Cref{a2-band-below-null}). n/a marks a quantity not measured for that model.}\tabularnewline
\toprule\noalign{}
\begin{minipage}[b]{\linewidth}\raggedright
Model
\end{minipage} & \begin{minipage}[b]{\linewidth}\raggedright
Decision-site PR, raw {[}95\% CI{]}
\end{minipage} & \begin{minipage}[b]{\linewidth}\raggedright
Decision-site PR, standardized
\end{minipage} & \begin{minipage}[b]{\linewidth}\raggedright
Content-position PR (last / mean)
\end{minipage} & \begin{minipage}[b]{\linewidth}\raggedright
Fraction of shuffle reference
\end{minipage} & \begin{minipage}[b]{\linewidth}\raggedright
Geometric-cell PR (raw $\to$ std)
\end{minipage} \\
\midrule\noalign{}
\endfirsthead
\toprule\noalign{}
\begin{minipage}[b]{\linewidth}\raggedright
Model
\end{minipage} & \begin{minipage}[b]{\linewidth}\raggedright
Decision-site PR, raw {[}95\% CI{]}
\end{minipage} & \begin{minipage}[b]{\linewidth}\raggedright
Decision-site PR, standardized
\end{minipage} & \begin{minipage}[b]{\linewidth}\raggedright
Content-position PR (last / mean)
\end{minipage} & \begin{minipage}[b]{\linewidth}\raggedright
Fraction of shuffle reference
\end{minipage} & \begin{minipage}[b]{\linewidth}\raggedright
Geometric-cell PR (raw $\to$ std)
\end{minipage} \\
\midrule\noalign{}
\endhead
\bottomrule\noalign{}
\endlastfoot
OLMo-3-7B-Instruct & 14.7 {[}14.3, 16.2{]} & 20.3 & 62.8 / 40.4 & 0.066 & 43 $\to$ 94 \\
Qwen2.5-7B & 8.6 {[}8.2, 9.4{]} & 13.5 & 42.4 / 32.7 & 0.041 & 1.0 $\to$ 39 \\
Llama-3.1-8B & 10.2 of record; 10.3 {[}10.1, 11.1{]} on this sample & 14.2 & 97.3 / 26.7 & 0.050 & 1.5 $\to$ 89 \\
GPT-OSS-20B & 9.4 {[}9.1, 10.7{]} & 12.8 & n/a & 0.084 & n/a \\
\end{longtable}

\subsection{Band-below-null means the position is invalid, not that the direction is absent}\label{a2-band-below-null}

\textbf{Failure as it appeared.} At the chat \path|final_pre_assistant| decision token on
OLMo-3-Instruct, the positive-control moral band came out at {[}0.40, 0.47{]}, and the honest
covariance-matched null came out at 0.557. Held-one-out moral directions projected onto
their own span \emph{below} where random directions projected. Read naively, any direction of
interest (refusal, judgment) that projected low there would read as ``not in the moral
subspace.'' That reading is unsupported: the instrument had no discriminating power at that
position, so it cannot certify absence of anything.

\textbf{The tell.} The positive control sits below the null. Band-below-null $\Rightarrow$ position-invalid
instrument. The moral band is not only a yardstick for ``moral-adjacent''; it is a validity
check on the measurement position. The cause here is dimensionality, not an outlier
dimension (at this decision token the top dim carries 0.2\% of variance) and not a null that standardization can
rescue (the null stays 0.52 after z-scoring). The channel is simply narrow: participation
ratio 14.7. The \texttt{$\surd$(3/14.7)\ =\ 0.45} heuristic (a rank-3 subspace at PR 14.7) predicts a
median-scale projection; comparing that 0.45 against a null q95 of 0.557 and against the
rank-3 pairwise-\textbar cos\textbar{} null of 0.41--0.51 is a consistency check that the numbers are the
right size, not a convergence of three independent estimates on one value (0.45 is a
median-scale prediction, 0.557 is a q95).

\textbf{The protocol.} \path|participation_ratio| is a required type-block field on every extracted
direction, and any position whose PR sits far below its own references is flagged
position-invalid for content projection-fraction tests at extraction time. The gate is stated
null-referenced rather than as an absolute number: the decision-site PR against the
column-shuffle reference (the PR the same marginals would give without correlations), against
the sample-rank ceiling \(n - 1\), and against the content positions of the same texts. All four
decision sites sit at 4 to 8 percent of the shuffle reference and 2.6 to 9.4 times below their
content positions (Table 1); an absolute threshold of 30 was the working rule in the program's
earlier sessions, and it is kept here only as the historical value. An absolute bar is not
evaluable at small sample sizes: with 64 rollouts the sample-rank ceiling is 63, and a content
position that reads 22 to 29 there is above its own covariance-matched sampling null while
failing the 30. (The false-invalid rate of the gate is not quantified beyond one panel case, GPT-OSS's
decision token, where the band survives the null despite a low PR; see Limitations.)

\textbf{The certifying check and the reframe.} Position-invalid does not mean uninterpretable
model. A projection-fraction test fails there, but a decision-\emph{direction} cosine does
not: it is immune to the projection null. In a \textasciitilde15-slot channel, refusal and judgment
directions occupy different slots at \textbar cos\textbar{} below even the low-dim random level, which reads
as active separation, not a weak-instrument artifact. Concretely, refusal-decision is
orthogonal to judgment-decision with no coupling detectable above \textbar cos\textbar{} 0.10 against a null
q95 of 0.41 on OLMo (0.32 vs 0.42 on Qwen, 0.08 vs 0.51 on Llama). Geometrically, the
moral-content band sits below the null at this bottleneck (band-below-null there, healthy at
content positions), so content-versus-decision orthogonality is structurally favored here.
This is a geometric observation, not a functional one: that moral content projects weakly
onto the decision channel does not by itself establish that it fails to reach the decision,
which is a causal claim that the note's own standard resolves only with an intervention cell.
Read as geometry, any comprehension-to-decision coupling would have to ride the attention
heads writing into the bottleneck, a concrete anatomical target.

One reconciling sentence is required for prose. The bottleneck is position-invalid for
content projection-fraction tests (band-below-null) and position-valid for decision-direction
reads (decision-direction cosine, and the GPT-OSS refusal projection). GPT-OSS's decision channel is called
``position-valid (PR 12.8)'' against a separate MoE PR sanity ceiling of 25; that ceiling is
not the content rule.

\begin{figure}[t]
\centering
\includegraphics[width=\linewidth]{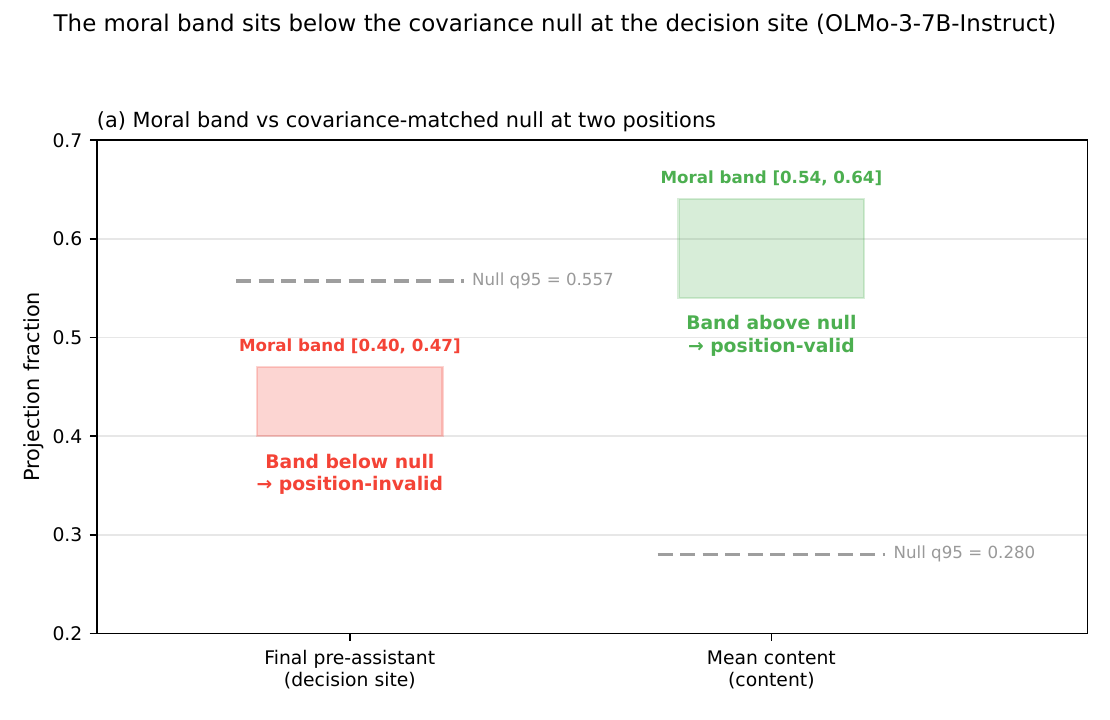}
\caption{The calibrated ladder at the OLMo-3-Instruct chat decision token. The positive-control moral band [0.40, 0.47] sits \emph{below} the covariance-matched null q95 of 0.557. A positive control below the null means the instrument has no discriminating power at that position for content projection-fraction tests: band-below-null implies the position is invalid, so a low projection there cannot certify absence. This is the visual form of the tell.}
\label{fig:ladder}
\end{figure}

\textbf{Figure~\ref{fig:ladder}} is the calibrated ladder at this position: the moral band {[}0.40, 0.47{]} plotted
below the covariance null 0.557, the visual form of the tell.

\subsection{The massive-activation outlier is position-dependent, so the bottleneck is clean}\label{a5-outlier}

\textbf{Failure as it appeared.} Llama-3.1 carries a massive-activation outlier: dim 788 holds
32\% of residual variance. The worry was that this outlier contaminated every geometric read
on Llama, including the decision-token cells.

\textbf{The tell.} The outlier's variance share is a \emph{content-position} statistic. The decision
token is a different position and had to be checked there, not assumed from the global
number.

\textbf{The protocol and check.} At the decision-token channel where the refusal and judgment
cells actually read, Llama is clean: participation ratio 13.5 (standardized, on that harness's
request-twin sample; 14.2 standardized and 10.2 raw on the in-format sample), covariance null 0.148, which
barely moves to 0.114 under per-dimension standardization. The outlier lives at content
positions, not at the \textasciitilde13-dim control-token decision bottleneck, which is clean and low-rank
across OLMo and Llama alike. So the ``decision site is a narrow control-token channel''
finding is cross-model, and the standardization fix matters more at content positions than at
the decision token. This is why the null degeneracy (next) and the bottleneck are two
different failures at two different positions, not one confound.

\subsection{Covariance-matched nulls degenerate in massive-activation families}\label{a1-covariance-null}

\textbf{Failure as it appeared.} The covariance-matched, rank-matched null (draw random
directions from \texttt{N(0,\ $\hat{\Sigma}$)} of residual activations, project onto the rank-r subspace) is the
honest null used throughout the program's earlier representational studies. On the instruct-model geometry it saturates:
the moral-subspace projection null q95 = 0.92 on Qwen and 0.36 on Llama, the pairwise-cosine
null q95 = 0.995 on Qwen and 0.90 on Llama, versus 0.26 on OLMo-3. At a saturated null every direction projects like a typical
direction, so the test has no discriminating power.

\textbf{The tell.} The null value itself is near its ceiling. The mechanism is the same massive
activations as the outlier finding above: Qwen dim 458 = 59\% of residual variance, Llama dim 788 = 32\%, OLMo-3's
top dim = 1.4\% at the content position of the geometric cell. \texttt{$\hat{\Sigma}$} is dominated by these dims, covariance-matched random directions nearly
all align with them, and they project \textasciitilde1 onto any subspace with a component there. The same
dims collapse distinct raw mean-diff directions (Qwen ethics $\approx$ moral mean-diff \textbar cos\textbar{} = 0.90).
This is the known massive-activations / attention-sink phenomenon \citep{sun2024massive,xiao2023efficient}.

\textbf{The protocol.} Recompute directions and the null in a per-dimension-standardized space
(z-score by $\sigma$ from a format/position-matched activation sample, sink tokens excluded), the
primary fix. The criterion-based robustness variant projects out each
dimension individually above 5\% of variance. Behavioral results (ablation, judgment
accuracy) never use this null and are untouched; only geometric cells need the re-audit.

\textbf{The certifying check.} The clean instrument must give the same verdict raw and
standardized: OLMo, whose activations are well-conditioned, does. The quantitative
before/after is the participation ratio (Table 1, geometric-cell column): raw PR = OLMo 43,
Qwen 1.0, Llama 1.5 (one dim carries essentially all variance for Qwen and Llama); after
z-scoring, PR = OLMo 94, Qwen 39, Llama 89. The raw PR $\approx$ 1 shows the collapse was near-total;
standardization lifts Qwen and Llama into a genuinely multi-dimensional space.

\textbf{A boundary case that names the residual limit.} On the refusal-projection cell the two
robustifications \emph{disagree}: standardization gives refusal 0.20 above controls 0.10
(strong-form false), while
top-k projection-out gives refusal 0.21 below controls 0.45--0.55 (strong-form true), and the
same split appears on Llama. When standardization and projection-out disagree, the subspace
is genuinely degenerate and needs a format or position change, not a null patch. The
in-format chat ladder (whose decision-site space carries no \textgreater5\%-variance dim, so it is
outlier-free by construction) is the discriminator. This is the entry's own thesis applied to
itself: no single null repair resolves a genuinely rank-1 space.

\textbf{Scope of the fix.} Which null a cell uses decides whether the degeneracy touches it. The
instruct-model moral-subspace projection cells use the covariance-matched null (the one that
degenerates); the program's Qwen/Llama geometric cells use a permutation test and raw
(unnormalized) projection fractions; the behavioral cells use no geometric null at all. We report which null each cell class uses;
on that accounting the permutation-and-raw-projection cells are not exposed to this degeneracy:
the permutation test's observed statistics are \textasciitilde0.01 (unsaturated), the raw projection fractions
are low and un-inflated (moral-subspace projection fraction 0.104 OLMo / 0.127 Qwen / 0.071
Llama, mean\textbar cos\textbar{} 0.04--0.07), and the moral-foundations subspace was built on the base model
whose foundation directions did not collapse onto the outlier dim. This ``not at risk'' reading
rests on a companion audit not released with this note and is not independently verifiable from
it. The degeneracy is confined
to the covariance-matched projection null applied to the instruct-model moral subspace. The
general caution stands: covariance-matched nulls silently degenerate in massive-activation
families, the field's default Llama/Qwen panel.

\subsection{Reordered-norm architectures overshoot naive per-head OV attribution \textasciitilde3\texorpdfstring{$\times$}{x}}\label{a3-ov-attribution}

\textbf{Failure as it appeared.} The Stage-1 write attribution on OLMo-3-7B-Instruct (sum of
per-head OV writes + per-layer MLP writes + embed onto the refusal direction, divided by the
true residual write at the read layer) came back at 3.05. The linear decomposition overshot
the actual residual write by 3$\times$. The original gate (\texttt{recon\ $\ge$\ 0.90}, one-sided) passed it,
because a floor only catches undershoot.

\textbf{The tell.} A reconstruction well above 1.0 on a decomposition that should sum to 1.0. The
mechanism is architectural: OLMo-2/3 use reordered (post-block) norm, applying
\path|post_attention_layernorm| to the attention output and \path|post_feedforward_layernorm| to the
MLP output \emph{before} the residual add, with no input norm. The true residual write of the
attention block is \texttt{RMSNorm($\Sigma$\_h\ W\_O\^{}h\ z\_h)}, not the raw sum; the naive OV decomposition
skips the norm, and since the raw block output has RMS above the norm's target it inflates
\textasciitilde3\texorpdfstring{$\times$}{x}. Pre-norm families (Llama, Qwen) write the raw block output to the residual and
reconstruct \textasciitilde1.0 natively, which is why the overshoot never appeared in Papers 1--7 (they used
activations and directions, never OV decomposition).

\textbf{The protocol.} A two-sided gate \texttt{0.90\ $\le$\ recon\ $\le$\ 1.10} (overshoot now fails), plus an exact
RMSNorm fold. Folding the block norm into per-head attribution is standard
interpretability tooling \citep{elhage2021framework, nanda2022transformerlens}; the contribution
this note claims is not the fold but quantifying the \textasciitilde3\texorpdfstring{$\times$}{x} overshoot it corrects on reordered-norm
architectures, plus the two-sided reconstruction gate that catches it (a one-sided floor misses
overshoot).
RMSNorm is diagonal at a fixed token, \texttt{norm(x)\ =\ ($\gamma$\ /\ rms(x))\ $\odot$\ x}, so
multiplying each pre-norm per-component write by the per-layer gain
\texttt{g\ =\ $\gamma$\ /\ sqrt(mean(x$^2$)\ +\ $\varepsilon$)} recovers the exact residual contribution. The fold fires
automatically for reordered-norm models (detected via \path|post_feedforward_layernorm|) and is a
no-op for pre-norm models.

\textbf{The certifying check.} The fold is exact: unit-tested to 1e-9, and it brings the Stage-1
reconstruction from 3.05 to 0.9999, inside the two-sided band. It affects only the head
anatomy on OLMo-2/3 and other reordered-norm families (the un-folded numbers are inflated,
for example the MLP write fraction was 0.23 un-folded and 0.384 folded). It does not touch the
decisive causal cell, which reads the model's real forward pass with no decomposition. Per-head
OV / logit-lens attribution silently overshoots \textasciitilde3\texorpdfstring{$\times$}{x} on reordered-norm models unless the block
norm is folded, a portable caution for a growing family (OLMo-2, OLMo-3, other post-norm
designs).

\section{Verdict discipline}\label{verdict-discipline}

Three estimator and intervention patterns gate how a number becomes a verdict.

\subsection{Ratio-of-ratios over MDE-crossing}\label{ratio-of-ratios}

Whether an effect clears its minimum detectable effect is power-dependent. Comparing two
effects by which side of the MDE each lands on is the overlap fallacy: it reads a difference
in power as a difference in kind. Compare two effects instead by a within-outcome ratio and a
bootstrap CI on the ratio difference.

\textbf{Worked case: the \path|under_transfer| reclassification.} The first headline was
\path|reads_non_vmoral_features| at n=11, resting on an absolute transport comparison (a
V\_moral-restricted patch clears its MDE, a comparison patch does not). That absolute
comparison was necessary but not sufficient. Re-run at n=23 with a within-outcome
normalization, the honest verdict was \path|under_transfer|: the restricted patch moves refusal
less than the full patch, but the two do not sit on opposite sides of a categorical line. The
powered decisive cells (n=23 request-twins) are full$\to$refusal $-$0.0833, V\_moral-restricted$\to$refusal
$-$0.0282, complement$\to$refusal $-$0.0636, harm-rank-1$\to$refusal $-$0.0261, random-rank-3$\to$refusal
$-$0.0005, full$\to$judgment +0.0459, restricted$\to$judgment +0.0237, against a refusal MDE of 0.0238
and a judgment MDE of 0.0086. \path|under_transfer| was then itself superseded by the rank sweep
(\path|harm_saturating|, \Cref{case-study}), but the estimator lesson is the one that recurs: the reclassification
from \path|reads_non_vmoral_features| to \path|under_transfer| happened because the ratio, not the
MDE-crossing, is the comparison of record. (The specificity claim that survives is stated as
a difference CI, not an overlap check: V\_moral-restricted moves refusal more than a random
rank-3, $\Delta$ = 0.031, paired 95\% CI {[}0.020, 0.043{]}, excludes 0.)

A second instance, from reasoning models: an early raw diff-of-means null (harm direction 0.44--0.49 of
residual norm) read as ``harmfulness is not causally encoded,'' but that was a magnitude
artifact. Reply-inversion \citep{zhao2025harmfulness} -- steering the model along the
harmfulness direction at the instruction token and measuring the fraction of replies that flip
between compliance and refusal -- produced a nonzero directional effect (Qwen2.5-14B-Instruct
shift +17.4 flips 33\%, Llama-3.1-8B-Instruct +3.0 flips 23\%). These are raw flip fractions with
no channel-matched random-direction specificity null at matched norm, so they establish ``a
directional intervention along the harm axis moves the reply,'' not that the harm axis does so
over any matched-norm direction. We ran that control afterwards on Llama-3.1-8B-Instruct (layer
12, 100 items, 20 random unit directions at the identical norm, half and full residual norm):
the harm direction flipped no reply at either norm and shifted every margin coherently toward
the safe side (mean shift \(-2.1\) and \(-2.6\) from a clean margin of \(-1.8\)), while random
matched-norm directions washed the margins toward zero and thereby flipped 61 and 83 percent
(q95) of the near-zero majority. The flip fraction is not a specific readout at that norm, and
the Llama reply-inversion number above does not reproduce under matched-norm steering; the
Qwen2.5-14B number was not re-tested. The specificity control is therefore not missing but
failed, which is the stronger statement of the same limitation. Magnitude and residual-norm
share are not causal relevance; a causal readout is.

\subsection{Power tables before compute}\label{power-tables}

Compute MDE(n) from measured within-condition variance before spending compute. If no
feasible n resolves the effect, the block is the instrument, not the sample, and the compute
run is futile.

\textbf{Worked case: the Llama bounded-unresolved table.} The Llama refusal cells came back chaotic
(\Cref{orthogonal-cell}, \Cref{stimulus-discipline}). The temptation was a larger same-design re-run. The power table, built from
saved within-level arrays, said the re-run was futile: the ratio-of-ratios CI on the latched
denominator was {[}$-$2.3, 4.9{]}, and no feasible n at that variance closes it, because the
denominator is saturated rather than noisy. The underpowered Llama \path|R_refusal_k| and
\texttt{R\_judgment\_\{k\textgreater{}1\}} cells were voided as denominator-latched, and the clean channel was
identified as the \emph{reverse} (disengage) direction, not more samples in the forward one. An
afternoon of saved-array work prevented a compute run. This is the general rule: saving per-pair /
per-rollout / per-head arrays by default keeps the power computation zero-GPU, so futility is
caught before the session, not after.

\subsection{The orthogonal-cell certificate}\label{orthogonal-cell}

When a causal readout comes back chaotic, root-split against an orthogonal outcome the same
intervention should move. If the orthogonal cell is coherent, the instrument is certified and
the chaos is a property of the read-out outcome, not a broken patch.

\textbf{Worked case: Llama content-swap.} On Llama the content-swap interchange patch produced
sign-chaotic refusal deltas (SD 0.31, median +0.029) against OLMo's clean $-$0.083, which read
as a broken instrument. The judgment cell is the positive control: the \emph{same} patch moved
judgment coherently (CI excludes 0). So the patch works. The refusal chaos is saturation: the
boundary-violating twins sit at the refusal ceiling (baseline refuse 0.83--1.0), so the
decision-token refusal projection is latched and has no room to move, and the refusal-delta SD
grows with severity (0.296 $\to$ 0.352) as saturation deepens. OLMo's refusal moved because it was
weak (unsaturated). A causal readout run at a saturated outcome yields chaotic, sign-unstable
deltas that mimic instrument failure; the orthogonal cell tells the two apart.

\section{Stimulus discipline}\label{stimulus-discipline}

The operating-point rule: discrimination screens must bracket the point where the outcome
actually moves. A saturated outcome latches the readout. Use a severity ladder and a boundary
band (outcome \textasciitilde0.5), and report the psychometric curve, not a single point.

This is where the readout itself can lie. Reasoning models defeat clean judgment readouts
(regex, final-answer, forced-logit), while a plain instruct model is clean on the same battery
(Qwen2.5-14B-Instruct: 24/24 harmless-safe, 24/24 harmful-harmful). The readout has to be
validated on the model class it is run on, at an operating point where the outcome is not
pinned.

\subsection{A deliberation/prefill asymmetry is operating-point-confounded when the gate is a step}\label{a6-deliberation}

\textbf{Failure as it appeared.} On GPT-OSS-20B the reasoning-prefill deliberation cell (engage =
inculpating prefill, disengage = exculpating prefill) emitted a clean-looking asymmetry
\texttt{A\ =\ 1.0}: engage flips benign$\to$refuse, disengage flips violating$\to$comply 0/7. It read as
one-way early commitment, as if deliberation only ever pushes toward refusal.

\textbf{The tell.} \texttt{A\ =\ 1.0} with a bootstrap CI of width 0. The disengage arm is uniformly 0, so
every resample returns 1: a degenerate CI, not a precise estimate. The rule-of-three exposes
it directly, disengage 0/7 gives a 95\% upper bound of \textasciitilde0.43, not 0. The asymmetry is the same
dynamic-range confound as \Cref{orthogonal-cell}: the disengage arm was tested on violating items that already
refuse at baseline (at the ceiling), while the engage arm was tested on unsaturated benign
items (with room to move up). An asymmetry statistic that compares an arm-with-headroom against
an arm-at-the-ceiling measures the operating points, not the model. A companion trap in the
same run: the harm-separability commitment curve reads \textasciitilde1.0 from the first trace bin, but that
measures when \emph{harm is represented} (harmful and harmless traces differ from the start),
not when the \emph{decision} is fixed.

\textbf{Why the usual fix is not enough here.} That fix was boundary-band twins (outcome \textasciitilde0.5).
GPT-OSS's gate is a step: the severity ladder finds no unsaturated violating level (empty
boundary band, 5.6\% of items in the mid-band). The operating point cannot be bracketed
behaviorally at the existing resolution, so more boundary-band stimuli do not exist to collect.

\textbf{The protocol (the de-confounder).} Replace the binary disengage flip with a graded
exculpatory prefill series (weak$\to$strong) and read a continuous projection (the decision-channel
residual under each prefill onto the refusal direction) alongside the behavioral flip. The
graded projection registers sub-flip movement, so ``no flip at maximum prefill'' splits cleanly
into \emph{reversible} (projection moves toward comply) versus \emph{genuine downward-robustness}
(projection flat). Saturation can no longer masquerade as commitment. A pre-registered
band-existence check decides whether a finer ladder is even buildable: the per-item base-refuse
histogram is read as smooth (resolution-limited, build a finer ladder) versus bimodal (a step,
switch to the graded readout).

\textbf{The certifying check.} These GPT-OSS behavioral cells are small, n = 7 to 10 items per
arm, and the flip fractions below should be read at that sample size. Under the graded series,
GPT-OSS is a reversible reader: strong exculpatory prefill flips ceiling-refusing violating
items to comply 6/10 (reported without a CI at n=10; 5/10 on a later replication of the same
items). The decision-channel refusal projection also moves monotonically toward comply along the
series, at the prefill token and at the post-response decision token, but a covariance-matched
random-direction null at each position shows the movement is not specific to the refusal direction
(one random direction in five moves as far; one-sided p 0.17 to 0.23), so the behavioral flip is the
only readout of record and the projection is reported as a position effect. The engage direction is
separately consequential: an inculpating-analysis prefill flips unsaturated benign requests to
refuse 7/7 (Wilson 95\% {[}0.65, 1.0{]}), so the decision is not fixed before the trace. The
first-run disengage 0/7 was the saturation trap, now resolved. Report the behavioral flip
and the graded readout separately; the asymmetry statistic itself is not reported, because it
is uninterpretable when one arm sits at the ceiling.

This is the operating-point rule's reasoning-model instance: when the gate is a step and the
boundary band is empty, do not report the asymmetry; switch to a graded intervention with a
continuous readout that registers sub-threshold movement.

\section{Depth discipline}\label{depth-discipline}

A verdict about what a circuit \emph{reads} must state the intervention depth relative to the
model's commitment. A patch at the read layer is post-commitment for an early-committing
model, so a ``reads-X'' or asymmetry claim measured there can be a read-layer artifact. Measure
at (and report) the pre-commitment coherent depth, and depth-match cross-model comparisons.

\textbf{Failure as it appeared.} The naive cross-model asymmetry at the read layer (layer 16) was
\texttt{A\_Llama\ =\ +0.82}, engage-dominant and latch-like, against \texttt{A\_OLMo\ =\ $-$0.20}, for a difference
of +1.03 (95\% CI {[}0.16, 1.61{]}, excludes 0). Read at face value, this said Llama's refusal has
a third distinct property, a hard directional latch that OLMo lacks.

\textbf{The tell.} Llama's disengage is coherent at earlier layers but not at the read layer.
Patch-layer sweep: Llama's disengage is coherent at layers 8/12/14 ($-$0.12 / $-$0.11 / $-$0.20,
CIs exclude 0) and incoherent at layer 16 ($-$0.014). Llama commits \emph{before} the read layer;
OLMo's disengage is coherent at layer 16 ($-$0.62), so OLMo commits at or after it. Measuring
Llama's asymmetry at layer 16 measures it after Llama has already decided, so the +0.82 is a
post-commitment artifact. (The two layer-12 disengage numbers are two cells: the patch-layer
sweep above reads $-$0.11, the depth-matched full re-run reads $-$0.57, both coherent; the verdict
is the same.)

\textbf{The protocol.} Depth-match the comparison to the pre-commitment coherent layer, layer 12,
and recompute both models there.

\textbf{The certifying check.} At matched layer 12, \texttt{A\_Llama\ =\ $-$0.28} (CI {[}$-$0.47, +0.03{]}) and
\texttt{A\_OLMo@12\ =\ $-$0.54} (CI {[}$-$0.81, $-$0.32{]}), a difference of +0.26, down from +1.03 at the read
layer. The Llama asymmetry CI {[}$-$0.47, +0.03{]} includes 0, and no difference-CI is reported on
the residual +0.26, so the surviving cross-model claim rests on the reads-axis difference, not
on a residual asymmetry. The apparent third property collapses: the asymmetry is a \emph{consequence}
of early-commitment, not an independent latch. What survives depth-matching is the reads-axis
difference: at layer 12 Llama reads broad (\path|broad_moral|: R\_refusal 0.85 $\approx$ R\_judgment 0.79,
gap closes, harm-rank-1 only 0.59) while OLMo stays harm-keyed (R\_refusal 0.43 \textless{} R\_judgment
0.53, gap open). The read-layer +0.82 is retained only as a voided number with its replacement,
never as a finding.

\begin{figure}[t]
\centering
\includegraphics[width=\linewidth]{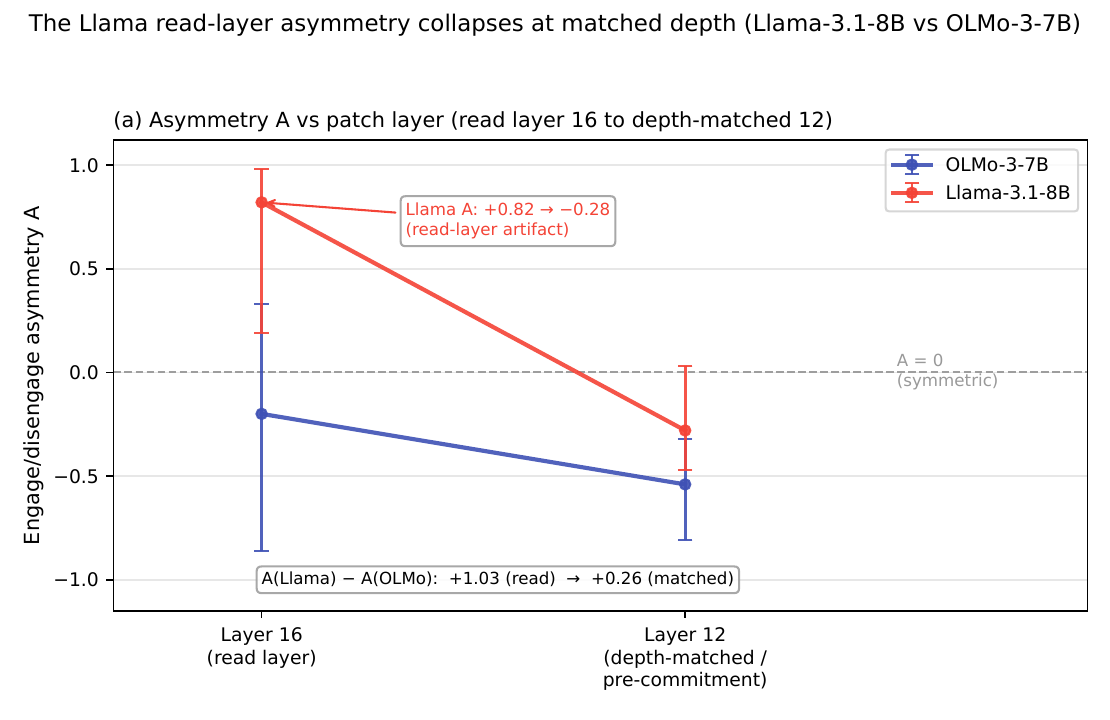}
\caption{Depth collapse of the cross-model asymmetry $A$. Measured at the read layer (16), $A_{\mathrm{Llama}}$ is $+0.82$ and reads as a Llama-specific directional latch. Depth-matched to the pre-commitment coherent layer (12), it falls to $-0.28$, while $A_{\mathrm{OLMo}}$ moves from $-0.20$ to $-0.54$; the cross-model difference collapses from $+1.03$ to $+0.26$. The apparent third property is a read-layer artifact of measuring after Llama has already committed.}
\label{fig:depth-collapse}
\end{figure}

\textbf{Figure~\ref{fig:depth-collapse}} is this depth collapse: \path|A_Llama| from +0.82 at the read layer to $-$0.28 at
matched layer 12, with \path|A_OLMo| $-$0.20 $\to$ $-$0.54, the depth-indexed exemplar.

\section{Case study: one program's amendments as caught failures}\label{case-study}

The program asked two questions of the refusal decision: \emph{what} moral content it
reads, and \emph{how} it commits. Its pre-registration trail is a sequence of caught
failures, each a dated amendment. Read in order, it is the methods note in miniature. The public amendment trail is a
credibility asset; it is cited from the flagship, not hidden.

\begin{itemize}
\tightlist
\item
  \textbf{Null degeneracy.} The instruct-model covariance null saturated on
  Qwen/Llama. Fix: standardized recompute; OLMo unchanged raw$\to$standardized certified it.
\item
  \textbf{Position gate.} The decision site is a PR-14.7 bottleneck with the
  positive control below the null. Fix: null-referenced PR position-invalid flag; V\_moral re-typed as
  format-robust (invalid-position artifact at \path|final_pre_assistant|, band matches at the valid
  \path|mean_content| position), and the content-projection numbers re-typed as non-verdict.
\item
  \textbf{Referee-pass hardening.} Before any asset was built, a referee pass
  re-typed the twin stimulus (request-twins carrying the judgment outcome, $\Delta$refusal expected-flat),
  added a transport positive control to the decisive cell, made the head-score null channel-matched
  (mean/resample ablation, not zeroing), and added a behavioral-discrimination pilot screen.
\item
  \textbf{OV overshoot.} Per-head attribution reconstructed at 3.05 on OLMo-3's
  reordered norm. Fix: two-sided gate + exact RMSNorm fold, reconstruction 3.05 $\to$ 0.9999.
\item
  \textbf{MDE-crossing headline.} The \path|reads_non_vmoral_features| verdict (n=11)
  rested on an absolute transport comparison. Fix: within-outcome ratio at n=23 $\to$ the honest
  \path|under_transfer|.
\item
  \textbf{Under-transfer superseded.} A rank sweep replaced the point comparison:
  as k $\in$ \{1, 3, 8, 16\}, R\_judgment climbs 0.05 $\to$ 0.46 $\to$ 0.59 $\to$ 0.66 while R\_refusal saturates
  0.01 $\to$ 0.31 $\to$ 0.26 $\to$ 0.27 at the harm-rank-1 level (harm\_rank1\_R 0.31), random-null \textasciitilde0 at
  every rank. The one-knob model \texttt{R\_refusal(k)\ $\approx$\ min(harm\_ceiling,\ R\_judgment(k))} fits the
  plateau (k$\ge$3) at RMSE 0.036, and PC1 (highest variance, purity 0.974, most harm-aligned at
  cos 0.35) is causally inert (rank-1 moves neither readout, 0.01 / 0.05), the lesson that
  variance is not causal relevance. Both the one-knob RMSE 0.036 and the PC1-inert reading are
  illustrative point estimates, reported without CIs at n=23. Verdict: \path|harm_saturating|.
\item
  \textbf{Replicated and pooled (W4, 2026-09).} A 19-twin replication through the same screen and
  harness reproduced the original subset (R\_refusal(16) 0.27, \path|harm_saturating|); the new twins
  alone read \texttt{indeterminate} with the same gap sign, and the pooled 42-twin sweep is primary:
  R\_refusal 0.03 $\to$ 0.27 $\to$ 0.22 $\to$ 0.24 {[}0.13, 0.41{]} against R\_judgment 0.05 $\to$ 0.46 $\to$ 0.59 $\to$ 0.66,
  one-knob ceiling 0.25 at RMSE 0.023, plateau interval a third narrower. The ratio-of-ratios
  secondary stayed unresolved (0.21, {[}$-$0.07, 0.39{]}), exactly as the pre-registered power table
  said it would at that count.
\item
  \textbf{GPT-OSS commit axis.} The first session banked the position gate (PR 12.8),
  consequential engage deliberation (benign$\to$refuse 7/7), and the first-run disengage 0/7 that
  looked irreversible.
\item
  \textbf{Power table.} The saved-array power computation ruled the Llama same-design
  re-run futile before the compute run (ratio-of-ratios CI {[}$-$2.3, 4.9{]} on a latched denominator).
\item
  \textbf{One-root diagnosis.} The Llama chaos was diagnosed by a single root split,
  judgment-delta coherence, not a grab-bag of probes: the orthogonal judgment cell is coherent,
  so the refusal chaos is saturation.
\item
  \textbf{Denominator-latched voids.} The Llama ``reads beyond harm'' hint (R\_refusal
  0.44 vs harm-rank-1 0.14 at rank 16) was voided, its denominator saturated; the three branches
  re-entered unweighted and were resolved by the depth-matched \path|broad_moral| read.
\item
  \textbf{Nomenclature + early-commitment.} Fixed engage = harm-add /
  disengage = harm-remove; defined the asymmetry statistic A; the patch-layer sweep gave the
  EARLY-COMMITMENT verdict (Llama disengage coherent at 8/12/14, incoherent at 16) and the
  read-layer cross-model asymmetry A\_Llama $-$ A\_OLMo = 1.03 (later depth-re-attributed, \Cref{depth-discipline}).
\item
  \textbf{Depth-indexed verdict.} The +0.82 read-layer asymmetry collapsed to $-$0.28
  at matched layer 12 (\Cref{depth-discipline}). This amendment started this note.
\item
  \textbf{Harm-coextensive hardening.} The reads-broad verdict survived the rank-1
  harm-coextensive alternative: a single harm cue spans only 3.6\% of the engage-driving moral
  basis, and the rank-2/4 severity-ladder version, run later, captures no more than a
  rank-matched sentiment basis (0.21 vs 0.24 at rank 4).
\item
  \textbf{Graded disengage.} The step-gate saturation trap was de-confounded: GPT-OSS
  is a reversible reader on behavior, violating$\to$comply 6/10 (5/10 on replication); the monotone
  projection that first corroborated it failed a later random-direction specificity null (\Cref{a6-deliberation}).
\item
  \textbf{Confound-named hypothesis.} The n=3 categorical co-occurrence
  (``harm-readers reversible, broad-reader early-commits'') was replaced by a falsifiable
  dimensionality$\to$reversibility hypothesis with an explicit architecture confound: the read$\leftrightarrow$commit
  pairing is confounded by lineage/scale/tokenizer/reasoning-vs-instruct at three points, and is
  deconfounded only by varying one axis at a time. The measured two-axis table stands; its
  interpretation is a follow-on hypothesis, not an n=3 claim.
\end{itemize}

\subsection{Reflexive discipline: the program audits its own published paper}\label{reflexive-discipline}

A cold-boot re-read turned the same scrutiny on the program's own published work. Paper 1 \citep[arXiv:2606.11375v1, 9 Jun 2026]{reblitzrichardson2026fragility} stated a raw layer-depth fragility gradient as its abstract-level Finding 2: late layers
were reported as far more fragile than early ones, with a raw late/early $\sigma$* ratio of 7 to 15$\times$ across
checkpoints (one checkpoint's late 10.0 / early 1.8 in that paper's Table 2), plus a raw post-saturation
$\sigma$* decline from 18.3 to 4.7. A post-submission control (\S4.4, RMS normalization) shows the
gradient is largely an activation-scale artifact: under RMS normalization the ratio collapses to
\textasciitilde1.8--2$\times$ (the residual \textasciitilde2$\times$ is not claimed as a genuine gradient, since RMS controls scale not
covariance shape), the cross-checkpoint ordering fails at 8/37 checkpoints, and the post-saturation
decline is withdrawn (flat, \textasciitilde13.8 $\to$ 15.0). The lesson is exact: raw $\sigma$* is valid within-layer (same activation
scale) but activation-scale-confounded cross-layer; RMS-normalize for any cross-layer claim.

Two things about \emph{how} it was caught belong in this note, and they are not the same thing.
First, what caught the error was a cold-boot (fresh-context) ledger re-read after publication,
not the calibration protocol firing at authoring time: the confound surfaced when a
fresh-context audit re-read a result the warm working sessions had produced and re-read many
times without flagging. The fresh-context reviewer's advantage is real, and mechanically
recreating it caught an abstract-level error. That the written calibration protocol would have
flagged this prospectively is a separate and weaker claim the note does not establish; the
honest record is that a published number stood until a fresh re-read caught it. Second, it
triggers a v2 erratum on a published paper. A program that runs an instrument-calibration discipline on other people's panels has to
run it on itself; the same scale confound named in the covariance null (magnitude is not
the signal) is the one that inflated Finding 2. This is the reflexive instance, and it is the
reason the note leads with ``instruments before verdicts'' rather than presenting the direction
results as settled.

\subsection{Claim hygiene: a traceable claim ledger}\label{claim-hygiene}

Every number in the program traces to an anchored-sentence row in a claim ledger: if a draft states
a scalar that ledger does not carry, the draft is wrong until a row is added. The ledger also
carries a register of superseded claims retained \emph{with their
replacements}, so they cannot re-enter prose as findings. The three the reader of a naive draft
would most likely resurrect are all there: the un-folded 3.05 head anatomy (replaced by the
folded 0.9999), the n=11 \path|reads_non_vmoral_features| headline (replaced by \path|under_transfer|
at n=23), and the \texttt{A\ =\ +0.82} read-layer asymmetry (replaced by the depth-matched $-$0.28).
A separate set of number-integrity flags blocks specific \emph{scalars}
whose value is still in dispute across documents, without blocking the verdicts (the shapes and
signs are robust; only a printed number waits on the flag). Voided results may be discussed as
methods lessons in this note; they are never findings in the flagship.

\section{Limitations}\label{limitations}

The discipline in this note is drawn from one research program, and its scope should be read
accordingly.

\textbf{Per-mode model coverage.} The summary ``six failures across four architectures'' is honest
only about the panel as a whole; each individual mode is established on one or two models, not
on all four. The band-below-null position-invalid instrument (\Cref{a2-band-below-null}) is shown on
OLMo-3-Instruct at its decision token, with the null-referenced PR gate applied on OLMo, Qwen, Llama, and GPT-OSS.
The massive-activation outlier's position-dependence (\Cref{a5-outlier}) is a Llama-3.1 finding cross-checked
against OLMo. The covariance-matched null degeneration (\Cref{a1-covariance-null}) is a Qwen-and-Llama result with
OLMo as the clean control. The reordered-norm OV overshoot (\Cref{a3-ov-attribution}) is OLMo-only, since pre-norm
Llama and Qwen reconstruct near 1.0 natively. The deliberation/prefill asymmetry (\Cref{a6-deliberation}) is
GPT-OSS-only. The read-layer depth artifact (\Cref{depth-discipline}) is Llama-versus-OLMo. So the note is six
protocols, each demonstrated on one or two members of a four-model panel, not six effects each
seen on four models.

\textbf{The position-validity gate can flag a real direction.} The PR gate declares a position
invalid for content projection-fraction tests, but its false-invalid rate is not quantified beyond
one panel case. A genuine content direction present at a narrow position would be flagged the same
way as a weak-instrument artifact, and on this panel that happens once in four: GPT-OSS's harmony
decision token sits at 8 percent of its shuffle reference and yet keeps its moral band above the
covariance null (0.53 against 0.48), so the PR gate alone would have flagged a position the band
test passes. The gate is calibrated to catch band-below-null cases; it is not
calibrated against a bank of known-present directions at narrow positions, so it can suppress a
real read. The gate is a validity flag for the projection instrument, not a statement that the position is
empty, and the band test, not the PR alone, is the deciding check.

\textbf{Standardization can destroy anisotropic signal.} Per-dimension standardization rescues the
covariance-matched null in massive-activation families (\Cref{a1-covariance-null}), but z-scoring flattens genuine
anisotropy along with the artifactual kind. The note's own boundary case, where standardization
and top-k projection-out disagree on the same cell (\Cref{a1-covariance-null}), is the symptom: when a subspace is
genuinely low-rank, standardization can report structure that projection-out does not, with no
guarantee the standardized space preserves a real anisotropic direction.

\textbf{Survivorship bias.} Only caught failures appear here. The note has no estimate of how many
measurement failures the discipline missed, and a protocol that reports only its catches cannot
state its own false-negative rate. The six are the ones a fresh-context re-read or a positive
control happened to surface; failures that produced a plausible number and were never
re-examined would not be in this note by construction.

\textbf{Single program, single panel.} All of the evidence is from one causal interpretability
program on one model panel (OLMo-3, Qwen2.5, Llama-3.1, GPT-OSS). The protocols are offered as
portable within that program and worth testing elsewhere, not as externally validated across
programs, tasks, or architectures beyond this panel.

\section{Checklist: the reusable protocol}\label{checklist}

The ship-blocker gates below are the portable form of the program's discipline. They are
the portable form of the program's internal checklists.

\textbf{Before any projection-fraction / cosine geometric cell:}

\begin{itemize}
\tightlist
\item
  Record \path|participation_ratio| at the measurement position with a subsampling interval, and
  state it against its references: the sample-rank ceiling \(n - 1\), the column-shuffle PR of the
  same marginals, and the content positions of the same texts. A position at a few percent of
  its shuffle reference is invalid for content projection-fraction tests; report a
  decision-direction cosine instead, or move to a valid position. (An absolute bar such as
  PR \textless{} 30 is not evaluable below \(n \approx 4 \cdot\)PR.)
\item
  Check the positive-control band against the covariance null at that position. Band-below-null
  means the instrument has no discriminating power there; do not report absence.
\item
  In a massive-activation family (any Llama/Qwen-class panel), inspect the null value for
  saturation and the top-dim variance share. Standardize (z-score, sinks excluded) and, as a
  robustness variant, project out each \textgreater5\%-variance dim. Certify with a clean model
  (raw$\to$standardized same verdict). When standardization and projection-out disagree, the space
  is genuinely degenerate; change format or position, not the null.
\end{itemize}

\textbf{Before any per-head OV / logit-lens attribution:}

\begin{itemize}
\tightlist
\item
  Detect reordered norm (post-block \path|post_feedforward_layernorm|). If present, fold the
  per-layer RMSNorm gain (unit-test the fold to \textasciitilde1e-9). Gate reconstruction two-sided
  (0.90 $\le$ recon $\le$ 1.10); a one-sided floor misses overshoot.
\end{itemize}

\textbf{Before any patch / ablation / interchange / steering verdict:}

\begin{itemize}
\tightlist
\item
  Pre-register the intervention spec block (stimulus--outcome baseline matching, transport
  positive control, channel-matched specificity null, token-alignment rule, harness parity for
  the outcome classifier) before the run.
\item
  Certify a chaotic readout with an orthogonal cell the same intervention should move; if the
  orthogonal cell is coherent, the chaos is saturation, not a broken instrument.
\item
  Compare two effects by a within-outcome ratio + bootstrap CI on the ratio difference, never
  by which side of the MDE each lands on.
\item
  State the intervention depth relative to commitment; depth-match cross-model comparisons to
  the pre-commitment coherent layer.
\end{itemize}

\textbf{Before any null / orthogonality / below-threshold verdict:}

\begin{itemize}
\tightlist
\item
  Build the calibrated ladder (floor $\to$ matched null $\to$ measurement $\to$ positive band) and re-verify
  every control's defining property in the current context.
\item
  State the minimum detectable effect. A null without an MDE has no teeth; write the detection
  bar into the sentence (``no coupling detectable above \textbar cos\textbar{} 0.10 against a null q95 of 0.41,''
  not ``dissociation'').
\end{itemize}

\textbf{Before any stimulus screen:}

\begin{itemize}
\tightlist
\item
  Bracket the operating point with a severity ladder and a boundary band (\textasciitilde0.5); report the
  psychometric curve. If the gate is a step (empty boundary band), switch to a graded
  intervention with a continuous readout and report the behavioral flip and the graded readout
  separately. Validate the readout on the model class it is run on.
\end{itemize}

\textbf{Before a compute run:}

\begin{itemize}
\tightlist
\item
  Compute MDE(n) from measured within-condition variance. If no feasible n resolves the effect,
  the block is the instrument, not the sample; do not spend the compute. Save per-pair /
  per-rollout / per-head arrays by default so the power computation and later statistics stay
  zero-GPU.
\end{itemize}

\textbf{Before any commit or draft:}

\begin{itemize}
\tightlist
\item
  Every printed scalar traces to an anchored claim-ledger row; voided numbers stay in a register
  of superseded claims with their replacements so they cannot re-enter prose as findings.
\end{itemize}

\section*{Reproducibility}\label{reproducibility}
\addcontentsline{toc}{section}{Reproducibility}

Each figure in this note ships with a regeneration script that reads a committed CSV under
the convention \path|papers/figure_data/mn_*.csv| (\path|mn_bottleneck_pr.csv|, \path|mn_ladder.csv|,
\path|mn_depth_collapse.csv|); the analysis outputs are gitignored, so the committed CSV plus its
script is the reproducibility contract for every figure.

Every number in this note, the participation-ratio profiles, the calibrated covariance nulls,
the positive-control ladders, the per-head write attribution, and the rank-sweep outcomes, is
indexed in the shared supplement \path|deepsteer/supplement/MANIFEST.json| (public repository:
\url{https://github.com/deepsteer/deepsteer/}), each with a content hash,
its provenance, and the figure or table it backs. The two instruments this note shares with the
companion flagship \citep{reblitzrichardson2026slice} (the decision-site participation-ratio profile and the depth-asymmetry panel)
live in the supplement once and are cited by both papers; \path|deepsteer/supplement/scripts/verify.py|
asserts the note's plotting copies match the canonical values, so a shared number can change in
only one place. Model ids, decision layers, standardization settings, and seeds are pinned in
\path|deepsteer/supplement/PROVENANCE.md|. Every array this note cites (the participation-ratio
samples with their bootstrap and null draws, the two-position validity ladder arrays, the
depth-matched per-twin deltas at layers 12 and 16, the RMSNorm-fold reconstruction arrays, and
the reply-inversion margins for the harm direction and twenty matched-norm random directions)
lives in the flagship's Zenodo deposit (\path|deepsteer_fl_arrays_v1.tar.zst|, CC BY 4.0, DOI
10.5281/zenodo.22731361); the deposit manifest lists the note's files by path
and SHA-256 so no shared array is duplicated.

\begin{ack}
This work made extensive use of Anthropic's Claude (the Claude Code agent on
Opus~4.6, 4.7, 4.8 and Fable~5) for code scaffolding, experimental scripts, and
prose drafting. The author retains responsibility for experimental design, all
scientific claims, and final wording.
\end{ack}

\bibliography{references}

\end{document}